\documentclass[preprint,12pt,3p]{elsarticle}
\usepackage{amssymb}
\usepackage[authoryear]{natbib}
\usepackage{hyperref}

\usepackage{todonotes}

\hypersetup{
    colorlinks=true,
    linkcolor=blue,
    filecolor=blue,      
    urlcolor=blue,
    citecolor=blue,
}
\journal{Elsevier}

\begin{document}
\begin{frontmatter}

\title{Monash Time Series Forecasting Archive}

\cortext[cor1]{Corresponding author. Postal Address: Faculty of Information Technology, P.O. Box 63 Monash University, Victoria
3800, Australia. E-mail address: rakshitha.godahewa@monash.edu}

\author[FIT]{Rakshitha Godahewa \corref{cor1}}

\author[FIT]{Christoph Bergmeir}

\author[FIT]{Geoffrey I.\ Webb}

\author[EBS]{Rob J. Hyndman}

\author[USYD]{Pablo Montero-Manso}

\address[FIT]{Department of Data Science and Artificial Intelligence, Monash University, Melbourne, Australia}
\address[EBS]{Department of Econometrics \& Business Statistics, Monash University, Melbourne, Australia}
\address[USYD]{University of Sydney, Australia}

\begin{abstract}
Many businesses and industries nowadays rely on large quantities of time series data making time series forecasting an important research area. Global forecasting models that are trained across sets of time series have shown a huge potential in providing accurate forecasts compared with the traditional univariate forecasting models that work on isolated series. However, there are currently no comprehensive time series archives for forecasting that contain datasets of time series from similar sources available for the research community to evaluate the performance of new global forecasting algorithms over a wide variety of datasets. In this paper, we present such a comprehensive time series forecasting archive containing 20 publicly available time series datasets from varied domains, with different characteristics in terms of frequency, series lengths, and inclusion of missing values. We also characterise the datasets, and identify similarities and differences among them, by conducting a feature analysis. Furthermore, we present the performance of a set of standard baseline forecasting methods over all datasets across eight error metrics, for the benefit of researchers using the archive to benchmark their forecasting algorithms. 
\end{abstract}

\end{frontmatter}

\section{Introduction}




Accurate time series forecasting is important for many businesses and industries to make decisions, and consequently, time series forecasting is a popular research area, lately in particular in machine learning.
A good benchmarking archive is essential for the growth of machine learning research \citep{tan_2020_monash}. Researchers gain the opportunity to perform a rigorous evaluation of their newly proposed machine learning algorithms over a large number of datasets when there is a proper benchmarking archive available. The University of California Irvine (UCI) repository \citep{dua_2017_uci} is the most common and well-known benchmarking archive used in general machine learning, and it has supported the development of many state-of-the-art algorithms. The UCI repository currently contains 507 datasets from various domains. 

In time series classification, there exist two well-known publicly available dataset archives: University of California Riverside (UCR) repository \citep{dau_2019_ucr} and University of East Anglia (UEA) repository \citep{bagnall_2018_uea}. The UCR repository was first released in 2002 with 16 datasets and at present, the number of datasets is 128, containing datasets of non-normalised time series, of varying lengths, and time series with missing values \citep{dau_2019_ucr}. The UCR repository contains only univariate time series datasets. Therefore, recently a group of researchers from UEA released the first multivariate time series classification archive which is known as UEA repository \citep{bagnall_2018_uea}. Currently, it contains 30 datasets containing equal length multivariate time series without any missing values. 
Furthermore, \citet{tan_2020_monash} recently released the first time series extrinsic regression archive containing 19 multi-dimensional time series datasets.

On the other hand, there are many time series forecasting competitions which have advanced the field. The most popular forecasting competition series is the M-competition series \citep{Makridakis_1982_TheAO, makridakis_2000_m3, makridakis_2018_m4, makridakis_2020_m4}. The latest competition of this series was the M5 competition \citep{makridakis_2020_m5} which is the fifth iteration of the series, and finished in July 2020. Other well-known forecasting competitions include the NN3 and NN5 Neural Network competitions \citep{BENTAIEB20127067}, the CIF 2016 competition \citep{burda_2017_cif}, and Kaggle competitions such as the Wikipedia web traffic \citep{google_2017_web} and Rossmann sales forecasting \citep{rossmann_2015_kaggle} competitions. 
The winning approaches of many of these competitions such as the winning method of the M4 forecasting by \citet{smyl_2020_hybrid} and the winning method of the M5 forecasting competition based on gradient boosted trees, consist of global forecasting models \citep{januschowski_2020_criteria} which train a single model across all series that need to be forecast. Compared with local models, global forecasting models have the ability to learn cross-series information during model training and can control model complexity and overfitting on a global level~\citep{pablo_2020_principles}. 
By now, many works have shown the superior performance of global models over per-series models in many situations \citep{bandara_2020_forecasting, hewamalage_2019_recurrent, pablo_2020_principles, flunkert_2017_deepar, treparo_2015_identification, godahewa2020weekly, godahewa2020ensembles, godahewa_2020_simulation}.

This can be seen as a paradigm shift in forecasting. Over decades, single time series were seen as a dataset that should be studied and modelled in isolation. Nowadays, we are oftentimes interested in global models built on sets of series from similar sources, such as series which are all product sales from a particular store, or series which are all smart meter readings in a particular city or state. 
Here, time series are seen as an instance in a dataset of many time series, to be studied and modelled together.


Though in the time series forecasting space there are a number of benchmarking archives, they follow the paradigm of single series as datasets, and consequently contain mostly unrelated single time series such as the Time Series Data Library \citep{hyndman_2018_tsdl}, ForeDeCk \citep{foredeck_2019}, the datasets of the M3 and M4 forecasting competitions, and others. These archives are of limited use to evaluate the performance of global forecasting models that typically are designed for, and perform better on, sets of series from similar sources (a notable exception from the rule being here ES-RNN by \citet{smyl_2020_hybrid}, the winning method of the M4 competition). There are currently no comprehensive time series forecasting benchmarking archives, to the best of our knowledge, that focus on such datasets of related time series to evaluate the performance of such global forecasting algorithms. Our aim is to address this limitation by  introducing such an archive. In particular, our paper has the following main contributions.

\begin{itemize}
    \item We introduce the first comprehensive time series forecasting archive containing datasets of related time series, available at \url{https://forecastingdata.org/}. This archive contains 20 publicly available time series datasets, with both equal and variable lengths time series. 
%
%
    Many datasets have different versions based on the frequency and the inclusion of missing values, making the total number of dataset variations to 50. Furthermore, it includes both real-world and competition time series datasets covering varied domains.
    \item We introduce a new format to store time series data, based on the Weka ARFF file format~\citep{arff_2008_paynter} and overcoming some of the shortcomings we observe in the .ts format used in the sktime time series repository \citep{sktime_2019_markus}. We use a .tsf extension for this new format. This format stores the meta-information about a particular time series dataset such as dataset name, frequency and inclusion of missing values as well as the series specific information such as starting timestamps, in a non-redundant way. The format is very flexible and capable of including any other attributes related to time series as preferred by the users.
    \item We analyse the characteristics of different series to identify the similarities and differences among them. For that, we  conduct a feature analysis using tsfeatures \citep{hyndman_2020_tsfeatures} and catch22 features \citep{lubba_2019_catch22} extracted from all series of all datasets. The extracted features are publicly available for further research use.
    \item We evaluate the performance of a set of baseline forecasting models including both traditional univariate forecasting models and global forecasting models over all datasets across eight error metrics. The forecasts and evaluation results of the baseline methods are publicly available for the benefits of researchers that use the repository to benchmark their forecasting algorithms.
    \item Finally, all implementations related to the forecasting archive including code for loading the datasets into the R and Python environments and code for feature calculations and evaluation of baseline forecasting models are publicly available at: \url{https://github.com/rakshitha123/TSForecasting}.
\end{itemize}

We also encourage other researchers to contribute time series datasets to our repository either by directly uploading them into the archive and/or by contacting the authors via email. 
The remainder of this paper is organised as follows: Section \ref{sec:datasets}  explains the datasets in the archive. Section \ref{sec:feature_analysis} explains the details of the feature analysis. Section \ref{sec:baseline} presents results of the baseline forecasting models over the datasets across eight error metrics. Finally, Section \ref{sec:conclusion} concludes the paper.



\section{Datasets}
\label{sec:datasets}

This section details the datasets in our time series forecasting archive. The current archive contains 20 time series datasets. Furthermore, the archive contains in addition 6 single very long time series. As a large amount of data oftentimes renders machine learning methods feasible compared with traditional statistical modelling, and we are not aware of good and systematic benchmark data in this space either, these series are included in our repository as well. A summary of all primary datasets included in the repository is shown in Table \ref{tab:datasets_summary}. 

%
A total of 50 datasets have been derived from these 26 primary datasets. Nine datasets contain time series belonging to different frequencies and the archive contains a separate dataset per each frequency. Seven of the datasets have series with missing values. The archive contains 2 versions of each of these, one with and one without missing values. In the latter case, the missing values have been replaced by using an appropriate imputation technique.

\begin{table}[htb]
\footnotesize
\begin{center}
\begin{tabular}{ l|l|l|c|c|c|c|c|c } 
\hline
& \textbf{Dataset} & \textbf{Domain} & \textbf{No: of} & \textbf{Min.} & \textbf{Max.} & \textbf{No: of} & \textbf{Missing} & \textbf{Competition}\\ 
&  &  & \textbf{Series} & \textbf{Length} & \textbf{Length} & \textbf{Freq.} & & \\ 
\hline
1 & M1 & Multiple & 1001 & 15 & 150 & 3 & No & Yes\\
2 & M3 & Multiple & 3003 & 20 & 144 & 4 & No & Yes\\
3 & M4 & Multiple & 100000 & 19 & 9933 & 6 & No & Yes\\
4 & Tourism & Tourism & 1311 & 11 & 333 & 3 & No & Yes\\
5 & NN5 & Banking & 111 & 791 & 791 & 2 & Yes & Yes\\
6 & CIF 2016 & Banking  & 72 & 34 & 120 & 1 & No & Yes\\
7 & Web Traffic  &  Web & 145063 & 803 & 803 & 2 & Yes & Yes\\
8 & Solar & Energy & 137 & 52560 & 52560 & 2 & No & No\\
9 & Electricity  & Energy & 321 & 26304 & 26304 & 2 & No & No\\
10 & London Smart Meters  & Energy & 5560 & 288 & 39648 & 1 & Yes & No\\
11 & Wind Farms & Energy & 339 & 6345 & 527040 & 1 & Yes & No \\
12 & Car Parts & Sales & 2674 & 51 & 51 & 1 & Yes & No\\
13 & Dominick & Sales & 115704 & 28 & 393 & 1 & No & No \\
14 & FRED-MD & Economic & 107 & 728 & 728 & 1 & No & No \\
15 & San Francisco Traffic & Transport & 862 & 17544 & 17544 & 2 & No & No\\
16 & Pedestrian Counts  & Transport & 66 & 576 & 96424 & 1 & No & No\\
17 & Hospital & Health & 767 & 84 & 84 & 1 & No & No \\
18 & COVID Deaths & Nature & 266 & 212 & 212 & 1 & No & No \\
19 & KDD Cup  & Nature & 270 & 9504 & 10920 & 1 & Yes & Yes\\
20 & Weather & Nature & 3010 & 1332 &  65981 & 1 & No & No \\
\hline
21 & Sunspot  & Nature & 1 & 73931 & 73931 & 1 & Yes & No\\
22 & Saugeen River Flow & Nature & 1 & 23741 & 23741 & 1 & No & No\\
23 & US Births & Nature & 1 & 7305 & 7305 & 1 & No & No\\
24 & Electricity Demand & Energy & 1 & 17520 & 17520 & 1 & No & No\\
25 & Solar Power & Energy & 1 & 7397222 & 7397222 & 1 & No & No\\
26 & Wind Power & Energy & 1 & 7397147 & 7397147 & 1 & No & No\\
\hline
\end{tabular}
\caption{Datasets in the Current Time Series Forecasting Archive}
\label{tab:datasets_summary}
\end{center}
\end{table}

Out of the 26 datasets, 8 originate from competition platforms, 3 from \citet{lai_2017_modeling}, 6 are taken from R packages, 1 is from the Kaggle platform \citep{kaggle_2019}, and 1 is taken from a Johns Hopkins repository \citep{CSSEGISandData_2020_repo}. The other datasets have been extracted from corresponding domain specific platforms. The datasets mainly belong to 9 different domains: tourism, banking, web, energy, sales, economics, transportation, health, and nature. Three datasets, the M1 \citep{Makridakis_1982_TheAO}, M3 \citep{makridakis_2000_m3}, and M4 \citep{makridakis_2018_m4, makridakis_2020_m4} datasets, contain series belonging to multiple domains.

For all datasets, we  introduce their key characteristics as well as the citations of the prior work where they have been used in the literature. Further details of these datasets are explained in the next sections, after describing the data format in which the series are stored. 


\subsection{Data Format}

\begin{figure}[htb]
\centering
  \includegraphics[scale=0.55]{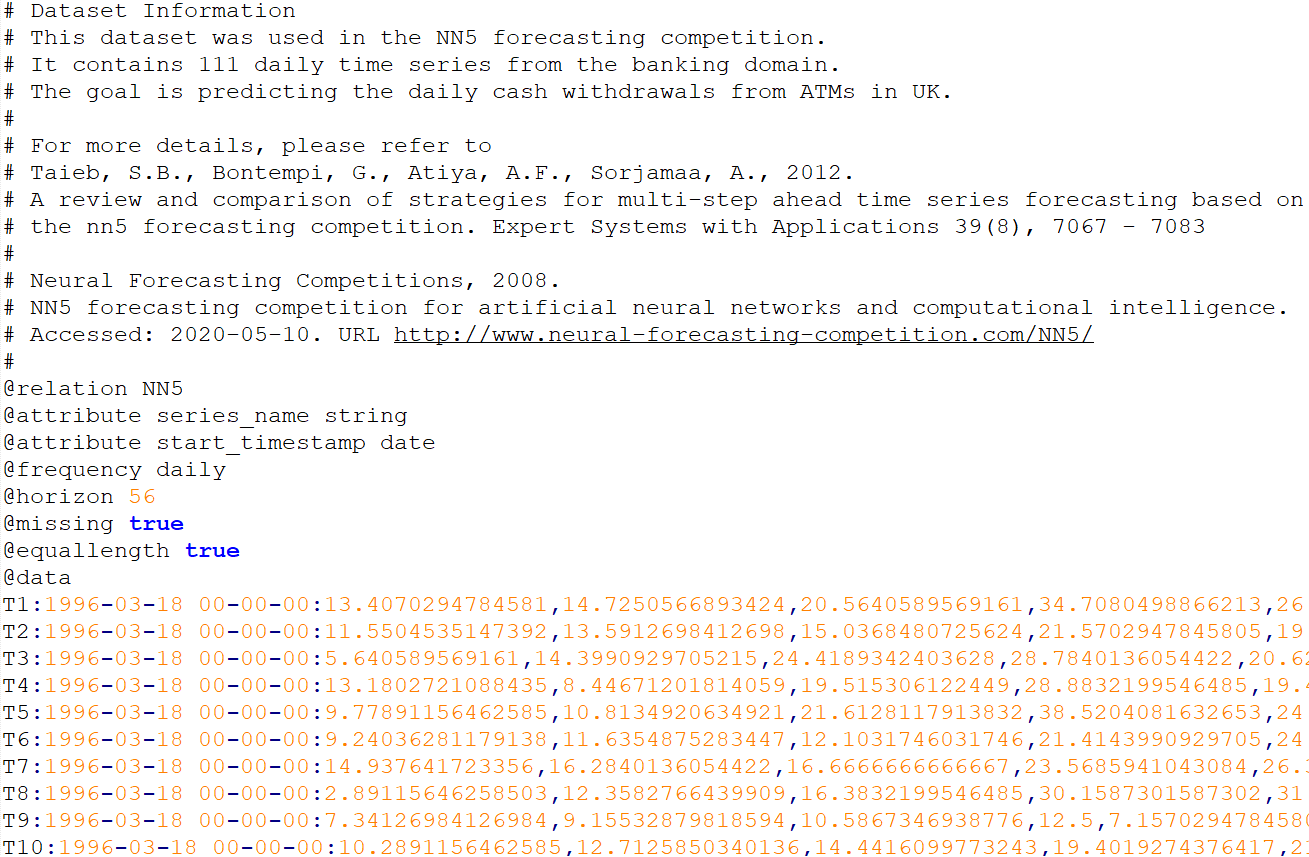}
  \caption{An example of the file format for the NN5 daily dataset}~\label{fig:data_format}
\end{figure}

We also introduce a new format to store time series data, based on the Weka ARFF file format~\citep{arff_2008_paynter}. We use the file extension .tsf and it is comparable with the .ts format used in the sktime time series repository \citep{sktime_2019_markus}, but we deem it more streamlined and more flexible.
The basic idea of the file format is that each data file can contain 1) attributes that are constant throughout the whole dataset (e.g., the forecasting horizon, whether the dataset contains missing values or not), 2) attributes that are constant throughout a time series (e.g., its name, its position in a hierarchy, product information for product sales time series), and 3) attributes that are particular to each data point (the value of the series, or timestamps for non-equally spaced series). An example of series in this format is shown in Figure~\ref{fig:data_format}.

The original Weka ARFF file format already deals well with the first two types of such attributes. Using this file format, in our format, each time series file contains tags describing the meta-information of the corresponding dataset such as \textit{@frequency} (seasonality), \textit{@horizon} (expected forecast horizon), \textit{@missing} (whether the series contain missing values) and \textit{@equallength} (whether the series have equal lengths). 
We note that these attributes can be freely defined by the user and the file format does not need any of these values to be defined in a certain way, though the file readers reading the files may rely on existence of attributes with certain names and assume certain meanings.
Next, there are attributes in each dataset which describe series-wise properties, where the tag \textit{@attribute} is followed by the name and type. Examples are \textit{series\_name} (the unique identifier of a given series) and \textit{start\_ timestamp} (the start timestamp of a given series). Again, the format has the flexibility to include any additional series-wise attributes as preferred by users. 

Following the ARFF file format, the data are then listed under the \textit{@data} tag after defining attributes and meta-headers, and attribute values are separated by colons. 
The only extension that our format has compared with the original ARFF file format, is that the time series then are appended to their attribute vector as a comma-separated variable-length vector. As this vector can have a different length for each instance, this cannot be represented in the original ARFF file format. In particular, a time series with $m$ number of attributes and $n$ number of values can be shown as:

\begin{equation}
\label{eqn:series_format}
    <attribute_1>:<attribute_2>:...:<attribute_m>:<s_1,s_2,...,s_n> 
\end{equation}

The missing values in the series are indicated using the ``?" symbol. Code to load datasets in this format into R and Python is available in our github repository.\footnote{\url{https://github.com/rakshitha123/TSForecasting}}


\subsection{Time Series Datasets}
This section describes the benchmark datasets that have a sufficient number of series from a particular frequency. The datasets may contain different categories in terms of domain and frequency.

\subsubsection{M1 Dataset}
The M1 competition dataset \citep{Makridakis_1982_TheAO} contains 1001 time series with 3 different frequencies: yearly, quarterly, and monthly as shown in Table \ref{tab:m1_dataset}. The series belong to 7 different domains: macro 1, macro 2, micro 1, micro 2, micro 3, industry, and demographic.

\begin{table}[htb]
\small
\begin{center}
\begin{tabular}{ l|c|c|c|c } 
\hline
\textbf{Frequency} & \textbf{No: of Series} & \textbf{Min. Length} & \textbf{Max. Length} & \textbf{Forecast Horizon} \\ 
\hline
Yearly & 181 & 15 & 58 & 6 \\
Quarterly & 203 & 18 & 114 & 8 \\
Monthly & 617 & 48 & 150 & 18 \\
\hline
Total & 1001 \\
\hline
\end{tabular}
\caption{Summary of M1 Dataset}
\label{tab:m1_dataset}
\end{center}
\end{table}

Research work which uses this dataset includes:
\begin{itemize}
    \item Forecasting with artificial neural networks: the state of the art \citep{zhang_1998_forecasting}
    \item Time series forecasting using a hybrid ARIMA and neural network model \citep{zhang_2003_time}
    \item Automatic time series forecasting: the forecast package for R \citep{hyndman_2008_automatic}
    \item Exponential Smoothing: the state of the art \citep{gardner_1985_exponential}
    \item Neural network forecasting for seasonal and trend time series \citep{zhang_2005_neural}
\end{itemize}

\subsubsection{M3 Dataset}
The M3 competition dataset \citep{makridakis_2000_m3} contains 3003 time series of various frequencies including yearly, quarterly, and monthly, as shown in Table \ref{tab:m3_dataset}. The series belong to 6 different domains: demographic, micro, macro, industry, finance, and other.

\begin{table}[htb]
\small
\begin{center}
\begin{tabular}{ l|c|c|c|c } 
\hline
\textbf{Frequency} & \textbf{No: of Series} & \textbf{Min. Length} & \textbf{Max. Length} & \textbf{Forecast Horizon}\\ \hline
Yearly &  645 & 20 & 47 & 6 \\
Quarterly & 756 & 24 & 72 & 8 \\
Monthly & 1428 & 66 & 144 & 18 \\
Other & 174 & 71 & 104 & 8 \\
\hline
Total & 3003 \\
\hline
\end{tabular}
\caption{Summary of M3 Dataset}
\label{tab:m3_dataset}
\end{center}
\end{table}

Research work which uses this dataset includes:
\begin{itemize}
    \item The theta model: a decomposition approach to forecasting \citep{assimakopoulos_2000_theta}
    \item Recurrent neural networks for time series forecasting: current status and future directions \citep{hewamalage_2019_recurrent}
    \item Out-of-sample tests of forecasting accuracy: an analysis and review \citep{TASHMAN2000437}
    \item Metrics for evaluating performance of prognostic techniques \citep{saxena_2008_metrics}
    \item Temporal link prediction using matrix and tensor factorizations \citep{dunlavy_2011_temporal}
    \item Forecasting time series with complex seasonal patterns using exponential smoothing \citep{livera_2011_forecasting}
    \item Evaluating forecasting methods \citep{armstrong_2001_evaluating}
    \item Exponential smoothing with a damped multiplicative trend \citep{taylor_2003_exponential}
\end{itemize}

\subsubsection{M4 Dataset}
The M4 competition dataset \citep{makridakis_2018_m4, makridakis_2020_m4} contains 100,000 time series with 6 different frequencies: yearly, quarterly, monthly, weekly, daily, and hourly, as shown in Table \ref{tab:m4_dataset}. The series belong to 6 different domains: demographic, micro, macro, industry, finance, and other, similar to the M3 forecasting competition. This dataset contains a subset of series available at ForeDeCk~\citep{foredeck_2019}.

\begin{table}[htb]
\small
\begin{center}
\begin{tabular}{ l|c|c|c|c } 
\hline
\textbf{Frequency} & \textbf{No: of Series} & \textbf{Min. Length} & \textbf{Max. Length} & \textbf{Forecast Horizon}\\ \hline
Yearly &  23000 & 19 & 841 & 6 \\
Quarterly &  24000  & 24 & 874  & 8 \\
Monthly &  48000  & 60 & 2812 & 18 \\
Weekly &  359  & 93 & 2610 & 13 \\
Daily & 4227  & 107 & 9933 & 14 \\
Hourly & 414  & 748 & 1008 & 48 \\
\hline
Total & 100000 \\
\hline
\end{tabular}
\caption{Summary of M4 Dataset}
\label{tab:m4_dataset}
\end{center}
\end{table}

Research work which uses this dataset includes:

\begin{itemize}
    \item A hybrid method of exponential smoothing and recurrent neural networks for time series forecasting \citep{smyl_2020_hybrid}
    \item FFORMA: Feature-based Forecast Model Averaging \citep{pablo_2020_fforma}
    \item Recurrent neural networks for time series forecasting: current status and future directions \citep{hewamalage_2019_recurrent}
    \item LSTM-MSNet: leveraging forecasts on sets of related time series with multiple seasonal patterns \citep{bandara_2019_msnet}
    \item Are forecasting competitions data representative of the reality? \citep{SPILIOTIS202037}
    \item Averaging probability forecasts: back to the future \citep{Winkler2018AveragingPF}
    \item A strong baseline for weekly time series forecasting \citep{godahewa2020weekly}
\end{itemize}

\subsubsection{Tourism Dataset}
This dataset originates from a Kaggle competition \citep{Athanasopoulos_2011_tourism, ellis_2018_tcomp} and contains 1311 tourism related time series with 3 different frequencies: yearly, quarterly, and monthly as shown in Table \ref{tab:tourism_dataset}.

\begin{table}[htb]
\small
\begin{center}
\begin{tabular}{ l|c|c|c|c } 
\hline
\textbf{Frequency} & \textbf{No: of Series} & \textbf{Min. Length} & \textbf{Max. Length} & \textbf{Forecast Horizon} \\ 
\hline
Yearly & 518 & 11 & 47 & 4 \\
Quarterly &  427 & 30 & 130 & 8 \\
Monthly & 366 & 91 & 333 & 24\\
\hline
Total & 1311 \\
\hline
\end{tabular}
\caption{Summary of Tourism Dataset}
\label{tab:tourism_dataset}
\end{center}
\end{table}

Research work which uses this dataset includes:
\begin{itemize}
    \item Recurrent neural networks for time series forecasting: current status and future directions \citep{hewamalage_2019_recurrent}
    \item A meta-analysis of international tourism demand forecasting and implications for practice \citep{PENG2014181}
    \item Improving forecasting by estimating time series structural components across multiple frequencies \citep{KOURENTZES2014291}
    \item Forecasting tourist arrivals using time-varying parameter structural time series models \citep{SONG2011855}
    \item Forecasting monthly and quarterly time series using STL decomposition \citep{THEODOSIOU20111178}
    \item A novel approach to model selection in tourism demand modeling \citep{AKIN201564}
\end{itemize}

\subsubsection{NN5 Dataset}
This dataset contains 111 time series of daily cash withdrawals from Automated Teller Machines (ATM) in the UK, and was used in the NN5 forecasting competition \citep{BENTAIEB20127067}. The forecast horizon considered in the competition was 56.
The original dataset contains missing values. 
Our repository contains two versions of the dataset: the original version with missing values and a modified version where the missing values have been replaced using a median substitution where a missing value on a particular day is replaced by the median across all the same days of the week along the whole series as in \citet{hewamalage_2019_recurrent}. 
Furthermore, \citet{godahewa2020weekly} use the weekly aggregated version of this dataset for their experiments related to proposing a baseline model for weekly forecasting. The aggregated weekly version of this dataset is also available in our repository.
Research work which uses this dataset includes:

\begin{itemize}
    \item Recurrent neural networks for time series forecasting: current status and future directions \citep{hewamalage_2019_recurrent}
    \item A strong baseline for weekly time series forecasting \citep{godahewa2020weekly}
    \item Forecasting across time series databases using recurrent neural networks on groups of similar series: a clustering approach \citep{bandara_2020_forecasting}
    \item Forecast combinations of computational intelligence and linear models for the NN5 time series forecasting competition \citep{ANDRAWIS2011672}
    \item Forecasting the NN5 time series with hybrid models \citep{WICHARD2011700}
    \item Multiple-output modeling for multi-step-ahead time series forecasting \citep{BENTAIEB20101950}
    \item Recursive multi-step time Series forecasting by perturbing data \citep{taieb_2011_recursive}
    \item Benchmarking of classical and machine-learning algorithms (with special emphasis on bagging and boosting approaches) for time series forecasting \citep{pritzsche_2015_benchmarking}
\end{itemize}



\subsubsection{CIF 2016 Dataset}
The dataset from the Computational Intelligence in Forecasting (CIF) 2016 forecasting competition contains 72 monthly time series.
Out of those, 24 series originate from the banking sector, and the remaining 48 series are artificially generated. There are 2 forecast horizons considered in the competition where 57 series have a forecasting horizon of 12 and the remaining 15 series consider the forecast horizon as 6 \citep{burda_2017_cif}. 
Research work which uses this dataset includes:
\begin{itemize}
    \item Recurrent neural networks for time series forecasting: current status and future directions \citep{hewamalage_2019_recurrent}
    \item Forecasting across time series databases using recurrent neural networks on groups of similar series: a clustering approach \citep{bandara_2020_forecasting}
    \item Improving time series forecasting: an approach combining bootstrap aggregation, clusters and exponential smoothing \citep{DANTAS2018748}
    \item Time series clustering using numerical and fuzzy representations \citep{afanasieva_2017_time}
    \item An automatic calibration framework applied on a metaheuristic fuzzy model for the CIF competition \citep{coelho_2016_automatic}
\end{itemize}

\subsubsection{Kaggle Web Traffic Dataset}

This dataset contains 145063 daily time series representing the number of hits  or web traffic for a set of Wikipedia pages from 01/07/2015 to 10/09/2017 used by the Kaggle web traffic forecasting competition \citep{google_2017_web}. The forecast horizon considered in the competition was 59.
As the original dataset contains missing values, we include both the original dataset in our repository and an imputed version. This dataset is intermittent and hence, we impute missing values with zeros.
Furthermore, \citet{godahewa2020weekly} use the weekly aggregated version of this dataset containing the first 1000 series. Our repository also contains this aggregated weekly version of the dataset for all series. The missing values of the original dataset were imputed before the aggregation.
Research work which uses this dataset includes:
\begin{itemize}
    \item Recurrent neural networks for time series forecasting: current status and future directions \citep{hewamalage_2019_recurrent}
    \item A strong baseline for weekly time series forecasting \citep{godahewa2020weekly}
    \item Web traffic prediction of Wikipedia pages \citep{petluri_2018_web}
    \item Improving time series forecasting using mathematical and deep learning models \citep{gupta_2018_improving}
    \item Foundations of sequence-to-sequence modeling for time series \citep{pmlr-v89-mariet19a}
\end{itemize}

\subsubsection{Solar Dataset}
This dataset contains 137 time series representing the solar power production recorded every 10 minutes in the state of Alabama in 2006. It was used by \citet{lai_2017_modeling}, and originally extracted from \citet{solar_dataset}.
Furthermore, \citet{godahewa2020weekly} use an aggregated version of this dataset containing weekly solar power production records. The aggregated weekly version of this dataset is also available in our repository.

\subsubsection{Electricity Dataset}
This dataset represents the hourly electricity consumption of 321 clients from 2012 to 2014 in kilowatt (kW). It was used by \citet{lai_2017_modeling}, and originally extracted from \citet{electricity_dataset}.
Our repository also contains an aggregated version of this dataset representing the weekly electricity consumption values.

\subsubsection{London Smart Meters Dataset} 
This dataset contains 5560 half-hourly time series that represent the energy consumption readings of London households in kWh from November 2011 to February 2014 \citep{smart_meter_2019_kaggle}. 
The series are categorized into 112 blocks in the original dataset. The series in our repository are in the same order (from block 0 to block 111) as they are in the original dataset.
The original dataset contains missing values and we impute them using the last observation carried forward (LOCF) method. Our repository contains both versions: the original version with missing values and the modified version where the missing values have been replaced.
Research work which uses this dataset includes:

\begin{itemize}
    \item Predicting electricity consumption using deep Recurrent Neural Networks \citep{Nugaliyadde2019PredictingEC}
    \item A single scalable LSTM model for short-term forecasting of disaggregated electricity loads \citep{Alonso2019ASS}
    \item Deep learning based short-term load forecasting for urban areas \citep{maksut_2019_deep}
    \item Smart grid energy management using RNN-LSTM: a deep learning-based approach \citep{kaur_2019_smart}
\end{itemize}

\subsubsection{Wind Farms Dataset}
This dataset contains very long minutely time series representing the wind power production of 339 wind farms in Australia. It was extracted from the Australian Energy Market Operator (AEMO) online platform \citep{nemweb_2020_wind}. The series in this dataset range from 01/08/2019 to 31/07/2020. 
The original dataset contains missing values where some series contain missing data for more than seven consecutive days. Our repository contains both the original version of the dataset and a version where the missing values have been replaced by zeros.





\subsubsection{Car Parts Dataset}
This dataset contains 2674 intermittent monthly time series showing car parts sales from January 1998 to March 2002.
It was extracted from the R package expsmooth \citep{hyndman_2015_expsmooth}. The package contains this dataset as \textit{``carparts"}.
As the original dataset contains missing values, we include the original version of the dataset in the repository as well as a version where the missing values have been replaced with zeros, as the series are intermittent.
Research work which uses this dataset includes:
\begin{itemize}
    \item Principles and algorithms for forecasting groups of time series: locality and globality \citep{pablo_2020_principles}
\end{itemize}

\subsubsection{Dominick Dataset}
This dataset contains 115704 weekly time series representing the profit of individual stock keeping units (SKU) from a retailer. 

It was extracted from the Kilts Center, University of Chicago Booth School of Business online platform \citep{dominicks_dataset}. This platform also contains daily store-level sales data on more than 3500 products collected from  Dominick's Finer Foods, a large American retail chain in the Chicago area, for approximately 9 years. The data are provided in different categories such as customer counts, store-specific demographics and sales products.
Research work which uses this dataset includes:

\begin{itemize}
    \item Principles and algorithms for forecasting groups of time series: locality and globality \citep{pablo_2020_principles}
    \item The value of competitive information in forecasting FMCG retail product sales and the variable selection problem \citep{HUANG2014738}
    \item  Beer snobs do exist: estimation of beer demand by type \citep{daniel_2014_beer}
    \item Downsizing and supersizing: how changes in product attributes influence consumer preferences \citep{jami_2014_downsizing}
    \item Reference prices, costs, and nominal rigidities \citep{eichenbaum_2011_reference}
    \item Sales and monetary policy \citep{guimaraes_2011_sales}
\end{itemize}

\subsubsection{FRED-MD Dataset}
This dataset contains 107 monthly time series showing a set of macro-economic indicators from the Federal Reserve Bank \citep{McCracken_2016_fred} starting from 01/01/1959. It was extracted from the FRED-MD database. The series are differenced and log-transformed as suggested in the literature.
Research work which uses this dataset includes:
\begin{itemize}
    \item Principles and algorithms for forecasting groups of time series: locality and globality \citep{pablo_2020_principles}
\end{itemize}

\subsubsection{San Francisco Traffic Dataset}
This dataset contains 862 hourly time series showing the road occupancy rates on San Francisco Bay area freeways from 2015 to 2016. It was used by \citet{lai_2017_modeling}, and originally extracted from \citep{traffic_dataset}.
\citet{godahewa2020weekly} use a weekly aggregated version of this dataset, which is also available in our repository.

\subsubsection{Melbourne Pedestrian Counts Dataset}
This dataset contains hourly pedestrian counts captured from 66 sensors in Melbourne city starting from May 2009 \citep{pedestrian_2017}. The original data are updated on a monthly basis when the new observations become available. The dataset in our repository contains pedestrian counts up to 30/04/2020. 
Research work which uses this dataset includes:

\begin{itemize}
    \item Enhancing pedestrian mobility in smart cities using big data \citep{carter_2020_enhancing}
    \item Visualising Melbourne pedestrian count \citep{obie_2017_visualising}
    \item PedaViz: visualising hour-level pedestrian activity \citep{obie_1028_pedaviz}
\end{itemize}

\subsubsection{Hospital Dataset}
This dataset contains 767 monthly time series showing the patient counts related to medical products from January 2000 to December 2006. 
It was extracted from the R package expsmooth \citep{hyndman_2015_expsmooth}. The package contains this dataset as \textit{``hospital"}.
Research work which uses this dataset includes:
\begin{itemize}
    \item Principles and algorithms for forecasting groups of time series: locality and globality \citep{pablo_2020_principles}
\end{itemize}

\subsubsection{COVID Deaths Dataset}
This dataset contains 266 daily time series that represent the total COVID-19 deaths in a set of countries and states from 22/01/2020 to 20/08/2020. It was extracted from the Johns Hopkins repository \citep{CSSEGISandData_2020_repo, dong_2020_interactive}. The original data are updated on a daily basis when the new observations become available.

\subsubsection{KDD Cup 2018 Dataset}
This competition dataset contains long hourly time series representing the air quality levels in 59 stations in 2 cities, Beijing (35 stations) and London (24 stations) from 01/01/2017 to 31/03/2018 \citep{kdd_2018}. The dataset represents the air quality in multiple measurements such as $PM2.5$, $PM10$, $NO_{2}$, $CO$, $O_{3}$ and $SO_{2}$ levels. 

Our repository dataset contains 270 hourly time series which have been categorized using city, station name, and air quality measurement. 

As the original dataset contains missing values, we include both the original dataset and an imputed version in our repository. We impute leading missing values with zeros and the remaining missing values using the LOCF method.
Research work which uses this dataset includes:

\begin{itemize}
    \item AccuAir: winning solution to air quality prediction for KDD cup 2018 \citep{luo_2019_kdd}
\end{itemize}

\subsubsection{Weather Dataset}
This dataset contains 3010 daily time series representing the variations of four weather variables: rain, minimum temperature, maximum temperature and solar radiation, measured at weather stations in Australia. The series were extracted from the R package \textit{bomrang} \citep{sparks_2020_bomrang}.
Research work which uses this dataset includes:
\begin{itemize}
    \item Principles and algorithms for forecasting groups of time series: locality and globality \citep{pablo_2020_principles}
\end{itemize}

\subsection{Single Long Time Series Datasets}

This section describes the benchmark datasets which have single time series with a large amount of data points.

\subsubsection{Sunspot Dataset}
The original data source contains a single very long daily time series of sunspot numbers from 01/01/1818 until the present \citep{sunspot_2015}. 
Furthermore, it also contains monthly mean total sunspot numbers (starting from 1749), 13-month smoothed monthly total sunspot numbers (starting from 1749), yearly mean total sunspot numbers (starting from 1700), daily hemispheric sunspot numbers (starting from 1992), monthly mean hemispheric sunspot numbers (starting from 1992), 13-month smoothed monthly hemispheric sunspot numbers (starting from 1992), and yearly mean total sunspot numbers (starting from 1610). The original datasets are updated as new observations become available.

Our repository contains the single daily time series representing the sunspot numbers from 08/01/1818 to 31/05/2020. 
As the dataset contains missing values, we include an LOCF-imputed version besides it in the repository. 
Research work which uses this dataset includes:
\begin{itemize}
    \item Re-evaluation of predictive models in light of new data: sunspot number version 2.0 \citep{gkana_2016_evaluation}
    \item Correlation between sunspot number and ca II K emission index \citep{bertello_2016_correlation}
    \item Dynamics of sunspot series on time scales from days to years: correlation of sunspot births, variable lifetimes, and evolution of the high-frequency spectral component \citep{shapoval_2017_dynamics}
    \item Long term sunspot cycle phase coherence with periodic phase disruptions \citep{pease_2016_long}
\end{itemize}

\subsubsection{Saugeen River Flow Dataset}
This dataset contains a single very long time series representing the daily mean flow of the Saugeen River at Walkerton in cubic meters per second from 01/01/1915 to 31/12/1979. The length of this time series is 23,741.
It was extracted from the R package, \textit{deseasonalize} \citep{mcLeod_2013_deseasonalization}. The package contains this dataset as \textit{``SaugeenDay"}.

Research work which uses this dataset includes:

\begin{itemize}
    \item Telescope: an automatic feature extraction and transformation approach for time series forecasting on a level-playing field \citep{bauer_2020_telescope}
\end{itemize}

\subsubsection{US Births Dataset}
This dataset contains a single very long daily time series representing the number of births in the US from 01/01/1969 to 31/12/1988. The length of this time series is 7,305.
It was extracted from the R package, \textit{mosaicData} \citep{randall_2020_mosaicData}. The package contains this dataset as \textit{``Births"}. 
Research work which uses this dataset includes:

\begin{itemize}
    \item Telescope: an automatic feature extraction and transformation approach for time series forecasting on a level-playing field \citep{bauer_2020_telescope}
\end{itemize}

\subsubsection{Electricity Demand Dataset}
This dataset contains a single very long time series representing the half hourly electricity demand for Victoria, Australia in 2014. The length of this time series is 17,520. 
It was extracted from the R package, \textit{fpp2} \citep{hyndman_2018_fpp2}. The package contains this dataset as \textit{``elecdemand"}. The temperatures corresponding with each demand value are also available in the original dataset.
%
Research work which uses this dataset includes:

\begin{itemize}
    \item Telescope: an automatic feature extraction and transformation approach for time series forecasting on a level-playing field \citep{bauer_2020_telescope}
\end{itemize}

\subsubsection{Solar Power Dataset}
This dataset contains a single very long time series representing the solar power production of an Australian wind farm recorded per each 4 seconds starting from 01/08/2019. It was extracted from the AEMO online platform \citep{nemweb_2020_wind}. The length of this time series is 7,397,222.

\subsubsection{Wind Power Dataset}
This dataset contains a single very long time series representing the wind power production of an Australian wind farm recorded per each 4 seconds starting from 01/08/2019. It was extracted from the AEMO online platform \citep{nemweb_2020_wind}. The length of this time series is 7,397,147.

\section{Feature Analysis}
\label{sec:feature_analysis}

We characterise the datasets in our archive to analyse the similarities and differences between them, to gain a better understanding on where gaps in the repository may be and what type of data are prevalent in real-world applications. This may also help to select suitable forecasting methods for different types of datasets.
%
We analyse the characteristics of the datasets using the tsfeatures \citep{hyndman_2020_tsfeatures} and catch22 \citep{lubba_2019_catch22} feature extraction methods. All extracted features are publicly available for further research use\footnote{\url{https://drive.google.com/drive/folders/1S-0LL-GQknu0jibX5bQQBDXbiChjU0NZ?usp=sharing}}. Due to their large size, we have not been able to extract features from the London smart meters, wind farms, solar power, and wind power datasets, which is why we exclude them from this analysis.

We extract 42 features using the \textit{tsfeatures} function in the R package \textit{tsfeatures} \citep{hyndman_2020_tsfeatures} including mean, variance, autocorrelation features, seasonal features, entropy, crossing points, flat spots, lumpiness, non-linearity, stability, Holt-parameters, and features related to the Kwiatkowski–Phillips–Schmidt–Shin (KPSS) test \citep{baum_2018_kpss} and the Phillips–Perron (PP) test \citep{phillips_1986_testing}. For all series that have a frequencies greater than daily, we consider multi-seasonal frequencies when computing features. Therefore, the amount of features extracted is higher for multi-seasonal datasets as the seasonal features are individually calculated for each season presented in the series. Furthermore, if a series is short and does not contain two full seasonal cycles, we calculate the features assuming a non-seasonal series (i.e., setting its frequency to ``one'' for the feature extraction).
%
We use the \textit{catch22\_all} function in the R package \textit{catch22} \citep{lubba_2018_catch22_pkg} to extract the catch22 features from a given time series. The features are a subset of 22 features from the \textit{hctsa} package \citep{fulcher_2017_hctsa} which includes the implementations of over 7000 time series features. The computational cost of the catch22 features is low compared with all features implementated in the hctsa package.

In the following feature analysis, we consider 5 features, as suggested by \citet{BOJER2020}: first order
autocorrelation (ACF1), trend, entropy, seasonal strength, and the Box-Cox transformation parameter, lambda.  The \textit{BoxCox.lambda} function in the R package \textit{forecast} \citep{hyndman_2008_automatic} is used to extract the Box-Cox transformation parameter from each series, with default parameters. The other 4 features are extracted using \textit{tsfeatures}. Since this feature space contains 5 dimensions, to compare and visualise the features across multiple datasets, we reduce the feature dimensionality to 2 using Principal Component Analysis \citep[PCA, ][]{Jolliffe2011}. 


\begin{figure}[htb]
\centering
  \includegraphics{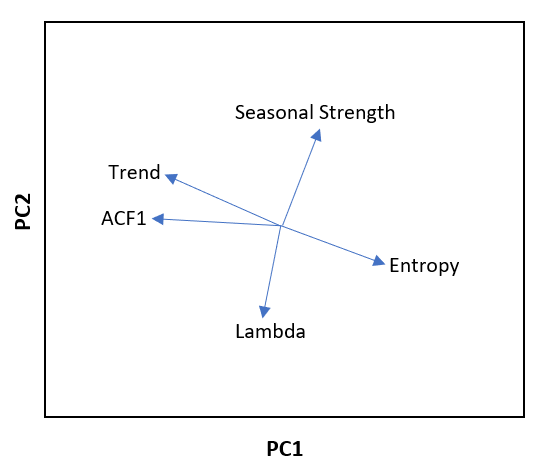}
  \caption{The directions of the 5 feature components: ACF1, trend, entropy, seasonal strength, and lambda, for the two-dimensional feature space generated by the first two principal components (PC1, PC2) extracted with PCA.}~\label{fig:components}
\end{figure}

The numbers of series in each dataset are significantly different, e.g., the CIF 2016 monthly dataset and M4 monthly dataset contain 72 and 48,000 series, respectively. Hence, if all series were considered to calculate the PCA components, those components would be dominated by datasets that have large amounts of series. Therefore, for datasets that contain more than 300 series, we randomly take a sample of 300 series, before constructing the PCA components across all datasets. Once the components are calculated, we map all series of all datasets into the resulting PCA feature space.
We note that we use PCA for dimensionality reduction over other advanced dimensionality reduction algorithms such as t-Distributed Stochastic Neighbor Embedding  \citep[t-SNE,][]{maaten_2008_visualizing} due to this capability of constructing the basis of the feature space with a reduced sample of series with the possibility to then map all series into the space afterwards. The directions of the 5 corresponding feature components generated by PCA are shown in Figure \ref{fig:components}. 

\begin{figure}
\centering
  \includegraphics[scale=0.46]{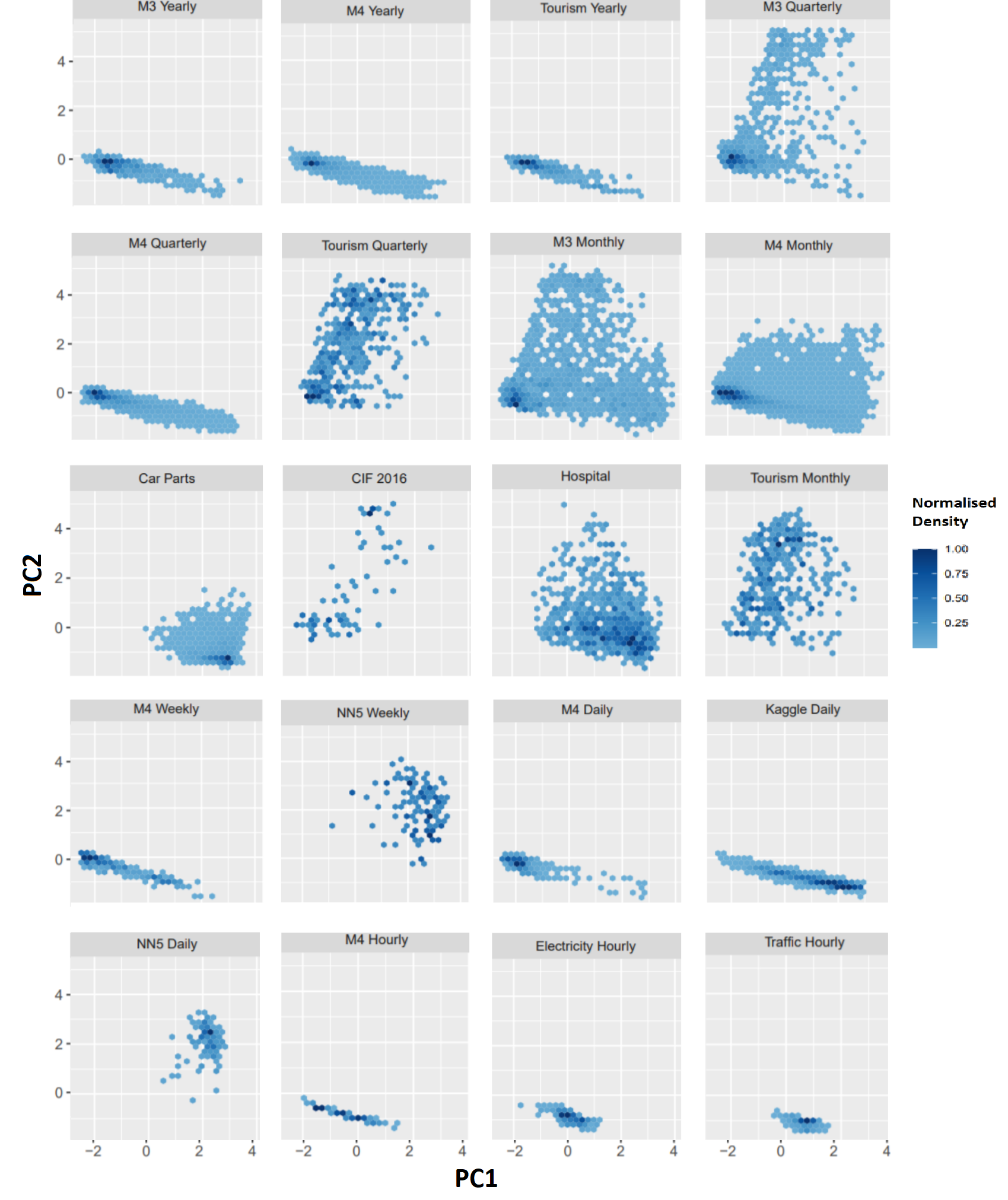}
  \caption{Hexbin plots showing the normalised density values of the low-dimensional feature space generated by PCA across ACF1, trend, entropy, seasonal strength and lambda for 20 datasets. The dark and light hexbins denote the high and low density areas, respectively.}~\label{fig:hexbin_tsfeatures}
\end{figure}

Figure \ref{fig:hexbin_tsfeatures} accordingly shows hexbin plots of the normalised density values for 20 datasets. The figure highlights the characteristics among different datasets. For the M competition datasets, the feature space is highly populated on the left-hand side and hence, denoting high trend and ACF1 levels in the series. The tourism yearly dataset also shows high trend and ACF1 levels. In contrast, the car parts, hospital, and Kaggle web traffic datasets show high density levels towards the right-hand side, indicating a higher degree of entropy. The presence of intermittent series can be considered the major reason for the higher degree of entropy in the Kaggle web traffic and car parts datasets. The plots confirm the claim by \citet{BOJER2020} and \citet{FRY2020156} that the M competition datasets are significantly different from the Kaggle web traffic dataset. 

The monthly datasets generally show high seasonal strengths compared with datasets of other frequencies. Quarterly datasets also demonstrate high seasonal strengths except for the M4 quarterly dataset. In contrast, the datasets with high frequencies such as weekly, daily, and hourly show low seasonal strengths except for the NN5 weekly and NN5 daily datasets. 

Related to the shapes of the feature space, the 3 yearly datasets: M3, M4, and tourism show very similar shapes and density populations indicating they have similar characteristics. The M4 quarterly dataset also shows a similar shape as the yearly datasets, even though it has a different frequency. The other 2 quarterly datasets M3 and tourism are different, but similar to each other. The M3 and M4 monthly datasets are similar to each other in terms of both shape and density population. Furthermore, the electricity hourly and traffic hourly datasets have similar shapes and density populations, whereas the M4 hourly dataset has a slightly different shape compared with them. The daily datasets show different shapes and density populations, where the NN5 daily dataset is considerably different from the other 2 daily datasets: M4 and Kaggle web traffic, in terms of shape and all 3 daily datasets are considerably different from each other in terms of density population. The weekly datasets also show different shapes and density populations compared with each other. 

PCA plots showing the normalized density values of all datasets corresponding with both tsfeatures and catch22 features are available in the Online Appendix\footnote{\url{https://drive.google.com/file/d/1qS6r-RV-Dac3Rj3Fv248o5hjnpZyEAIk/view?usp=sharing}}.

\section{Baseline Forecasting Models}
\label{sec:baseline}

In this section, we evaluate the performance of different baseline forecasting models including 6 traditional univariate forecasting models and a global forecasting model over the datasets in our repository using a fixed origin evaluation scheme, so that researchers that use the data in our repository can directly benchmark their forecasting algorithms against these baselines. The following 7 baseline forecasting methods are considered for the experiments:

\begin{itemize}
    \item Exponential Smoothing \citep[ETS,][]{hyndman_2008_ets}
    \item Auto-Regressive Integrated Moving Average \citep[ARIMA,][]{box_1990_arima}
    \item Simple Exponential Smoothing (SES)
    \item Theta \citep{assimakopoulos_2000_theta}
    \item Trigonometric Box-Cox ARMA Trend Seasonal \citep[TBATS,][]{livera_2011_forecasting}
    \item Dynamic Harmonic Regression ARIMA \citep[DHR-ARIMA,][]{hyndman_2018_fpp2}
    \item A globally trained Pooled Regression model \citep[PR,][]{treparo_2015_identification} 
\end{itemize}

Again, we do not consider the London smart meters, wind farms, solar power and wind power datasets for both univariate and PR model evaluations, and the Kaggle web traffic daily dataset for the PR model evaluation, as the computational cost of running these models was not feasible in our experimental environment. 

We use the R packages \textit{forecast} \citep{hyndman_2015_forecast} and \textit{glmnet} \citep{friedman_2010_glmnet} to implement the 6 traditional univariate forecasting methods and the globally trained PR method, respectively. 

The Theta, SES, and PR methods are evaluated for all datasets. ETS and ARIMA are evaluated for yearly, quarterly, monthly, and daily datasets. We consider the datasets with small frequencies, namely, 10 minutely, half hourly, and hourly as multi-seasonal and hence, TBATS and DHR-ARIMA are evaluated for those datasets instead of ETS and ARIMA due to their capability of dealing with multiple seasonalities \citep{bandara_2019_msnet}. TBATS and DHR-ARIMA are also evaluated for weekly datasets due to their capability of dealing with long non-integer seasonal cycles present in weekly data \citep{godahewa2020weekly}. 


Forecast horizons are chosen for each dataset to evaluate the model performance. For all competition datasets, we use the forecast horizons originally employed in the competitions. For the remaining datasets, 12 months ahead forecasts are obtained for monthly datasets, 8 weeks ahead forecasts are obtained for weekly datasets, except the solar weekly dataset, and 30 days ahead forecasts are obtained for daily datasets. For the solar weekly dataset, we use a horizon of 5 as the series in this dataset are relatively short compared with other weekly datasets. For half-hourly, hourly and other low frequency datasets, we set the forecasting horizon to one week, e.g., 168 is used as the horizon for hourly datasets.

The number of lagged values used in the PR models are determined similar to the heuristic suggested by \citet{hewamalage_2019_recurrent}. Generally, the number of lagged values is chosen as the seasonality multiplied with 1.25. If the datasets contain short series and it is impossible to use the above defined number of lags, for example in the Dominick and solar weekly datasets, then the number of lagged values is chosen as the forecast horizon multiplied with 1.25, assuming that the horizon is not arbitrarily chosen and reveals certain characteristics of the time series structure. When defining the number of lagged values for multi-seasonal datasets, we consider the corresponding weekly seasonality value, e.g., 168 for hourly datasets. If it is impossible to use the number of lagged values obtained with the weekly seasonality due to high memory and computational requirements, for example with the traffic hourly and electricity hourly datasets, then we use the corresponding daily seasonality value to define the number of lags, e.g., 24 for hourly datasets. In particular, due to high memory and computational requirements, the number of lagged values is chosen as 50 for the solar 10 minutely dataset which is less than the above mentioned heuristics based on seasonality and forecasting horizon suggest.  


We use four error metrics common for evaluation in forecasting, namely the Mean Absolute Scaled Error \citep[MASE,][]{hyndman_2006_another}, symmetric Mean Absolute Percentage Error (sMAPE), Mean Absolute Error \citep[MAE,][]{sammut_2010_mae}, and Root Mean Squared Error (RMSE). For datasets containing zeros, calculating the sMAPE error measure may lead to divisions by zero. Hence, we also consider the variant of the sMAPE proposed by \citet{suilin_2017_kaggle} which overcomes the problems with small values and divisions by zero of the original sMAPE. We report the original sMAPE only for datasets where divisions by zero do not occur. Equations \ref{eqn:mase}, \ref{eqn:smape}, \ref{eqn:msmape}, \ref{eqn:mae}, and \ref{eqn:rmse}, respectively, show the formulas of MASE, sMAPE, modified sMAPE, MAE, and RMSE, where $M$ is the number of data points in the training series, $S$ is the seasonality of the dataset, $h$ is the forecast horizon, $F_k$ are the generated forecasts and $Y_k$ are the actual values. We set the parameter $\epsilon$ in Equation \ref{eqn:msmape} to its proposed default of 0.1. 

\begin{equation}
\label{eqn:mase}
    MASE = \frac{\sum_{k=M+1}^{M+h} |F_{k} - Y_{k}|}{\frac{h}{M - S}\sum_{k=S+1}^{M} |Y_{k} - Y_{k - S}|} 
\end{equation}

\begin{equation}
\label{eqn:smape}
    sMAPE = \frac{100\%}{h}\sum_{k=1}^{h} \frac{|F_{k} - Y_{k}|}{(|Y_{k}| + |F_{k}|)/2} 
\end{equation}

\begin{equation}
\label{eqn:msmape}
    msMAPE = \frac{100\%}{h}\sum_{k=1}^{h} \frac{|F_{k} - Y_{k}|}{ max(|Y_{k}| + |F_{k}| + \epsilon, 0.5 + \epsilon)/2} 
\end{equation}

\begin{equation}
\label{eqn:mae}
    MAE = \frac{\sum_{k=1}^{h}{|F_{k} - Y_{k}|}}{h}
\end{equation}

\begin{equation}
\label{eqn:rmse}
    RMSE = \sqrt{\frac{\sum_{k=1}^{h}{|F_{k} - Y_{k}|}^2}{h}}
\end{equation}



The MASE measures the performance of a model compared with the in-sample average performance of a one-step-ahead na\"ive or sna\"ive benchmark. For multi-seasonal datasets, we use the lowest frequency to calculate the MASE. For the datasets where all series contain at least one full seasonal cycle of data points, we consider the series to be seasonal and calculate MASE values using the sna\"ive benchmark. Otherwise, we calculate the MASE using the na\"ive benchmark, effectively treating the series as non-seasonal.

The error metrics are defined for each series individually. We further calculate the mean and median values of the error metrics over the datasets to evaluate the model performance and hence, each model is evaluated using 10 error metrics for a particular dataset: mean MASE, median MASE, mean sMAPE, median sMAPE, mean msMAPE, median msMAPE, mean MAE, median MAE, mean RMSE and median RMSE. Tables \ref{tab:mean_mase}, \ref{tab:median_mase}, \ref{tab:mean_smape}, \ref{tab:median_smape}, \ref{tab:mean_msmape}, \ref{tab:median_msmape},  \ref{tab:mean_mae}, \ref{tab:median_mae}, \ref{tab:mean_rmse}, and \ref{tab:median_rmse} show the mean MASE, median MASE, mean sMAPE, median sMAPE, mean msMAPE, median msMAPE, mean MAE, median MAE, mean RMSE, and median RMSE results, respectively, of the SES, Theta, ETS, ARIMA, TBATS, DHR-ARIMA, and PR models on the same 20 datasets we considered for the feature analysis. The results of all baselines across all datasets are available in the Online Appendix.

Overall, SES shows the worst performance and Theta shows the second-worst performance across all error metrics. 
ETS and ARIMA show a mixed performance on the yearly, monthly, quarterly, and daily datasets but both outperform SES and Theta. 
TBATS generally shows a better performance than DHR-ARIMA on the high frequency datasets.  For our experiments, we always set the maximum order of Fourier terms used with DHR-ARIMA to $k = 1$. Based on the characteristics of the datasets, $k$ can be tuned as a hyperparameter and it may lead to better results compared with our results. Compared with SES and Theta, both TBATS and DHR-ARIMA show superior performance. 

\begin{table}
\begin{center}
\footnotesize
\begin{tabular}{lccccccc}
\hline
\textbf{Dataset} & \textbf{SES} & \textbf{Theta} & \textbf{ETS} & \textbf{ARIMA} & \textbf{TBATS} & \textbf{DHR-ARIMA} & \textbf{PR} \\ 
\hline
NN5 Daily & 1.521 & 0.885 & 0.865 & 1.013 & - & - & 1.263 \\ 
  NN5 Weekly & 0.903 & 0.885 & - & - & 0.872 & 0.887 & 0.854 \\ 
  CIF 2016 & 1.291 & 0.997 & 0.841 & 0.929 & - & - & 1.019 \\ 
  Kaggle Daily & 0.924 & 0.928 & 1.231 & 0.890 & - & - & - \\ 
  Tourism Yearly & 3.253 & 3.015 & 3.395 & 3.775 & - & - & 3.516 \\ 
  Tourism Quarterly & 3.210 & 1.661 & 1.592 & 1.782 & - & - & 1.643 \\ 
  Tourism Monthly & 3.306 & 1.649 & 1.526 & 1.589 & - & - & 1.678 \\ 
  Traffic Hourly & 1.922 & 1.922 & - & - & 2.482 & 2.535 & 1.281 \\ 
  Electricity Hourly & 4.544 & 4.545 & - & - & 3.690 & 4.602 & 2.912 \\ 
  M3 Yearly & 3.167 & 2.774 & 2.860 & 3.417 & - & - & 3.223 \\ 
  M3 Quarterly & 1.417 & 1.117 & 1.170 & 1.240 & - & - & 1.248 \\ 
  M3 Monthly & 1.091 & 0.864 & 0.865 & 0.873 & - & - & 1.010 \\ 
  M4 Yearly & 3.981 & 3.375 & 3.444 & 3.876 & - & - & 3.625 \\ 
  M4 Quarterly & 1.417 & 1.231 & 1.161 & 1.228 & - & - & 1.316 \\ 
  M4 Monthly & 1.150 & 0.970 & 0.948 & 0.962 & - & - & 1.080 \\ 
  M4 Weekly & 0.587 & 0.546 & - & - & 0.504 & 0.550 & 0.481 \\ 
  M4 Daily & 1.154 & 1.153 & 1.239 & 1.179 & - & - & 1.162 \\ 
  M4 Hourly & 11.607 & 11.524 & - & - & 2.663 & 13.557 & 1.662 \\ 
  Carparts & 0.897 & 0.914 & 0.925 & 0.926 & - & - & 0.755 \\ 
  Hospital & 0.813 & 0.761 & 0.765 & 0.787 & - & - & 0.782 \\ 
 \hline
\end{tabular}
\caption{Mean MASE Results}
\label{tab:mean_mase}
\end{center}
\end{table}

\begin{table}
\begin{center}
\footnotesize
\begin{tabular}{lccccccc}
\hline
\textbf{Dataset} & \textbf{SES} & \textbf{Theta} & \textbf{ETS} & \textbf{ARIMA} & \textbf{TBATS} & \textbf{DHR-ARIMA} & \textbf{PR} \\ 
\hline
NN5 Daily & 1.482 & 0.838 & 0.809 & 0.926 & - & - & 1.224 \\ 
  NN5 Weekly & 0.781 & 0.805 & - & - & 0.827 & 0.769 & 0.781 \\ 
  CIF 2016 & 0.862 & 0.662 & 0.532 & 0.559 & - & - & 0.746 \\ 
  Kaggle Daily & 0.539 & 0.548 & 0.667 & 0.528 & - & - & - \\ 
  Tourism Yearly & 2.442 & 2.360 & 2.373 & 2.719 & - & - & 2.356 \\ 
  Tourism Quarterly & 2.309 & 1.348 & 1.275 & 1.388 & - & - & 1.361 \\ 
  Tourism Monthly & 2.336 & 1.382 & 1.276 & 1.337 & - & - & 1.484 \\ 
  Traffic Hourly & 1.817 & 1.817 & - & - & 1.380 & 2.365 & 1.228 \\ 
  Electricity Hourly & 4.766 & 4.766 & - & - & 2.300 & 4.630 & 2.878 \\
  M3 Yearly & 2.261 & 1.985 & 1.907 & 2.003 & - & - & 2.267 \\ 
  M3 Quarterly & 1.073 & 0.831 & 0.855 & 0.917 & - & - & 0.902 \\ 
  M3 Monthly & 0.861 & 0.721 & 0.712 & 0.704 & - & - & 0.825 \\ 
  M4 Yearly & 2.940 & 2.312 & 2.329 & 2.753 & - & - & 2.568 \\ 
  M4 Quarterly & 1.142 & 0.973 & 0.886 & 0.925 & - & - & 1.038 \\ 
  M4 Monthly & 0.867 & 0.763 & 0.736 & 0.727 & - & - & 0.844 \\ 
  M4 Weekly & 0.441 & 0.416 & - & - & 0.365 & 0.382 & 0.392 \\ 
  M4 Daily & 0.862 & 0.861 & 0.859 & 0.867 & - & - & 0.868 \\ 
  M4 Hourly & 3.685 & 3.688 & - & - & 1.873 & 3.507 & 1.010 \\ 
  Carparts & 0.562 & 0.482 & 0.562 & 0.600 & - & - & 0.375 \\ 
  Hospital & 0.745 & 0.723 & 0.731 & 0.733 & - & - & 0.740 \\ 
 \hline
\end{tabular}
\caption{Median MASE Results}
\label{tab:median_mase}
\end{center}
\end{table}

\begin{table}
\begin{center}
\footnotesize
\begin{tabular}{lccccccc}
\hline
\textbf{Dataset} & \textbf{SES} & \textbf{Theta} & \textbf{ETS} & \textbf{ARIMA} & \textbf{TBATS} & \textbf{DHR-ARIMA} & \textbf{PR} \\ 
\hline
NN5 Daily & 35.50 & 22.01 & 21.57 & 26.01 & - & - & 30.30 \\ 
  NN5 Weekly & 12.24 & 11.96 & - & - & 11.63 & 11.84 & 11.45 \\ 
  CIF 2016 & 14.95 & 13.05 & 12.18 & 11.70 & - & - & 12.33 \\ 
  Kaggle Daily & - & - & - & - & - & - & - \\ 
  Tourism Yearly & 34.14 & 31.96 & 36.56 & 33.44 & - & - & 46.94 \\ 
  Tourism Quarterly & 27.41 & 15.37 & 15.07 & 16.58 & - & - & 15.86 \\ 
  Tourism Monthly & 36.39 & 19.90 & 19.02 & 19.73 & - & - & 21.11 \\ 
  Traffic Hourly & - & 82.44 & - & - & 70.59 & 92.58 & - \\ 
  Electricity Hourly & - & - & - & - & 40.47 & - & - \\ 
  M3 Yearly & 17.76 & 16.76 & 17.00 & 18.84 & - & - & 17.13 \\ 
  M3 Quarterly & 10.90 & 9.20 & 9.68 & 10.24 & - & - & 9.77 \\ 
  M3 Monthly & 16.22 & 13.86 & 14.14 & 14.24 & - & - & 15.17 \\ 
  M4 Yearly & 16.40 & 14.56 & 15.36 & 16.03 & - & - & 14.53 \\ 
  M4 Quarterly & 11.08 & 10.31 & 10.29 & 10.52 & - & - & 10.84 \\ 
  M4 Monthly & 14.38 & 13.01 & 13.53 & 13.08 & - & - & 13.74 \\ 
  M4 Weekly & 9.01 & 7.83 & - & - & 7.30 & 7.94 & 7.43 \\ 
  M4 Daily & 3.05 & 3.07 & 3.13 & 3.01 & - & - & 3.06 \\ 
  M4 Hourly & 42.95 & 42.98 & - & - & 28.12 & 35.99 & 11.68 \\ 
  Carparts & - & - & - & - & - & - & - \\ 
  Hospital & 17.98 & 17.31 & 17.50 & 17.83 & - & - & 17.60 \\ 
 \hline
\end{tabular}
\caption{Mean sMAPE Results}
\label{tab:mean_smape}
\end{center}
\end{table}

\begin{table}
\begin{center}
\footnotesize
\begin{tabular}{lccccccc}
\hline
\textbf{Dataset} & \textbf{SES} & \textbf{Theta} & \textbf{ETS} & \textbf{ARIMA} & \textbf{TBATS} & \textbf{DHR-ARIMA} & \textbf{PR} \\ 
\hline
NN5 Daily & 34.68 & 20.56 & 20.35 & 22.80 & - & - & 28.81 \\ 
  NN5 Weekly & 10.95 & 10.96 & - & - & 10.97 & 11.08 & 10.50 \\ 
  CIF 2016 & 11.40 & 7.95 & 6.58 & 7.69 & - & - & 8.43 \\ 
  Kaggle Daily & - & - & - & - & - & - & - \\ 
  Tourism Yearly & 18.81 & 16.83 & 19.20 & 22.66 & - & - & 16.88 \\ 
  Tourism Quarterly & 22.48 & 13.17 & 12.89 & 13.13 & - & - & 13.33 \\ 
  Tourism Monthly & 30.24 & 17.40 & 17.16 & 18.01 & - & - & 18.47 \\ 
  Traffic Hourly & - & 74.21 & - & - & 55.69 & 86.56 & - \\ 
  Electricity Hourly & - & - & - & - & 23.23 & - & - \\ 
  M3 Yearly & 12.44 & 11.54 & 11.52 & 12.37 & - & - & 12.92 \\ 
  M3 Quarterly & 6.74 & 5.23 & 5.53 & 6.36 & - & - & 5.73 \\ 
  M3 Monthly & 10.71 & 9.25 & 9.13 & 9.01 & - & - & 10.40 \\ 
  M4 Yearly & 11.41 & 9.23 & 8.97 & 10.20 & - & - & 9.49 \\ 
  M4 Quarterly & 6.94 & 6.06 & 5.61 & 5.80 & - & - & 6.34 \\ 
  M4 Monthly & 8.38 & 7.24 & 7.00 & 7.13 & - & - & 8.20 \\ 
  M4 Weekly & 5.17 & 5.19 & - & - & 4.81 & 5.10 & 4.99 \\ 
  M4 Daily & 1.99 & 2.01 & 1.99 & 2.01 & - & - & 2.00 \\ 
  M4 Hourly & 19.88 & 19.79 & - & - & 6.55 & 32.18 & 5.80 \\ 
  Carparts & - & - & - & - & - & - & - \\ 
  Hospital & 16.58 & 15.91 & 16.13 & 16.77 & - & - & 16.14 \\ 
 \hline
\end{tabular}
\caption{Median sMAPE Results}
\label{tab:median_smape}
\end{center}
\end{table}

\begin{table}
\begin{center}
\footnotesize
\begin{tabular}{lccccccc}
\hline
\textbf{Dataset} & \textbf{SES} & \textbf{Theta} & \textbf{ETS} & \textbf{ARIMA} & \textbf{TBATS} & \textbf{DHR-ARIMA} & \textbf{PR} \\ 
\hline
NN5 Daily & 35.38 & 21.93 & 21.49 & 25.91 & - & - & 30.20 \\ 
  NN5 Weekly & 12.24 & 11.96 & - & - & 11.62 & 11.83 & 11.45 \\ 
  CIF 2016 & 14.94 & 13.04 & 12.18 & 11.69 & - & - & 12.32 \\ 
  Kaggle Daily & 45.87 & 47.98 & 57.94 & 44.39 & - & - & - \\ 
  Tourism Yearly & 34.10 & 31.93 & 36.52 & 33.39 & - & - & 46.92 \\ 
  Tourism Quarterly & 27.41 & 15.37 & 15.07 & 16.58 & - & - & 15.86 \\ 
  Tourism Monthly & 36.39 & 19.89 & 19.02 & 19.73 & - & - & 21.11 \\ 
  Traffic Hourly & 8.73 & 8.73 & - & - & 12.58 & 11.72 & 5.97 \\ 
  Electricity Hourly & 44.39 & 44.94 & - & - & 40.15 & 43.78 & 30.00 \\ 
  M3 Yearly & 17.76 & 16.76 & 17.00 & 18.84 & - & - & 17.13 \\ 
  M3 Quarterly & 10.90 & 9.20 & 9.68 & 10.24 & - & - & 9.77 \\ 
  M3 Monthly & 16.22 & 13.86 & 14.14 & 14.24 & - & - & 15.17 \\ 
  M4 Yearly & 16.40 & 14.56 & 15.36 & 16.03 & - & - & 14.53 \\ 
  M4 Quarterly & 11.08 & 10.31 & 10.29 & 10.52 & - & - & 10.83 \\ 
  M4 Monthly & 14.38 & 13.01 & 13.52 & 13.08 & - & - & 13.73 \\ 
  M4 Weekly & 9.01 & 7.83 & - & - & 7.30 & 7.94 & 7.43 \\ 
  M4 Daily & 3.04 & 3.07 & 3.13 & 3.01 & - & - & 3.06 \\ 
  M4 Hourly & 42.92 & 42.94 & - & - & 28.10 & 35.94 & 11.67 \\ 
  Carparts & 64.88 & 59.27 & 65.76 & 65.61 & - & - & 43.23 \\ 
  Hospital & 17.94 & 17.27 & 17.46 & 17.79 & - & - & 17.56 \\ 
 \hline
\end{tabular}
\caption{Mean msMAPE Results}
\label{tab:mean_msmape}
\end{center}
\end{table}

\begin{table}
\begin{center}
\footnotesize
\begin{tabular}{lccccccc}
\hline
\textbf{Dataset} & \textbf{SES} & \textbf{Theta} & \textbf{ETS} & \textbf{ARIMA} & \textbf{TBATS} & \textbf{DHR-ARIMA} & \textbf{PR} \\ 
\hline
NN5 Daily & 34.57 & 20.51 & 20.31 & 22.72 & - & - & 28.70 \\ 
  NN5 Weekly & 10.94 & 10.96 & - & - & 10.97 & 11.08 & 10.50 \\ 
  CIF 2016 & 11.40 & 7.95 & 6.58 & 7.69 & - & - & 8.43 \\ 
  Kaggle Daily & 37.02 & 37.69 & 46.19 & 35.16 & - & - & - \\ 
  Tourism Yearly & 18.77 & 16.83 & 19.04 & 22.57 & - & - & 16.88 \\ 
  Tourism Quarterly & 22.48 & 13.17 & 12.89 & 13.13 & - & - & 13.33 \\ 
  Tourism Monthly & 30.24 & 17.40 & 17.16 & 18.00 & - & - & 18.47 \\ 
  Traffic Hourly & 8.26 & 8.26 & - & - & 7.58 & 10.66 & 5.43 \\ 
  Electricity Hourly & 42.08 & 42.20 & - & - & 23.22 & 38.30 & 24.78 \\ 
  M3 Yearly & 12.44 & 11.54 & 11.52 & 12.37 & - & - & 12.92 \\ 
  M3 Quarterly & 6.74 & 5.23 & 5.53 & 6.36 & - & - & 5.73 \\ 
  M3 Monthly & 10.71 & 9.25 & 9.13 & 9.01 & - & - & 10.39 \\ 
  M4 Yearly & 11.41 & 9.23 & 8.97 & 10.20 & - & - & 9.49 \\ 
  M4 Quarterly & 6.94 & 6.06 & 5.61 & 5.80 & - & - & 6.34 \\ 
  M4 Monthly & 8.38 & 7.24 & 7.00 & 7.13 & - & - & 8.20 \\ 
  M4 Weekly & 5.17 & 5.19 & - & - & 4.81 & 5.10 & 4.99 \\ 
  M4 Daily & 1.99 & 2.01 & 1.99 & 2.01 & - & - & 2.00 \\ 
  M4 Hourly & 19.86 & 19.75 & - & - & 6.55 & 32.08 & 5.80 \\ 
  Carparts & 45.45 & 45.45 & 46.18 & 46.18 & - & - & 30.30 \\ 
  Hospital & 16.57 & 15.90 & 16.11 & 16.75 & - & - & 16.12 \\ 
 \hline
\end{tabular}
\caption{Median msMAPE Results}
\label{tab:median_msmape}
\end{center}
\end{table}

\begin{table}
\begin{center}
\footnotesize
\begin{tabular}{lccccccc}
\hline
\textbf{Dataset} & \textbf{SES} & \textbf{Theta} & \textbf{ETS} & \textbf{ARIMA} & \textbf{TBATS} & \textbf{DHR-ARIMA} & \textbf{PR} \\ 
\hline
NN5 Daily & 6.63 & 3.80 & 3.72 & 4.41 & - & - & 5.47 \\ 
  NN5 Weekly & 15.66 & 15.30 & - & - & 14.98 & 15.38 & 14.94 \\ 
  CIF 2016 & 581875.97 & 714818.58 & 642421.42 & 469059.49 & - & - & 563205.57 \\ 
  Kaggle Daily & 363.43 & 358.73 & 403.23 & 340.36 & - & - & - \\ 
  Tourism Yearly & 95579.23 & 90653.60 & 94818.89 & 95033.24 & - & - & 82682.97 \\ 
  Tourism Quarterly & 15014.19 & 7656.49 & 8925.52 & 10475.47 & - & - & 9092.58 \\ 
  Tourism Monthly & 5302.10 & 2069.96 & 2004.51 & 2536.77 & - & - & 2187.28 \\ 
  Traffic Hourly & 0.03 & 0.03 & - & - & 0.04 & 0.04 & 0.02 \\ 
  Electricity Hourly & 845.97 & 846.03 & - & - & 574.30 & 868.20 & 537.38 \\ 
  M3 Yearly & 1022.27 & 957.40 & 1031.40 & 1416.31 & - & - & 1018.48 \\ 
  M3 Quarterly & 571.96 & 486.31 & 513.06 & 559.40 & - & - & 519.30 \\ 
  M3 Monthly & 743.41 & 623.71 & 626.46 & 654.80 & - & - & 692.97 \\ 
  M4 Yearly & 1009.06 & 890.51 & 920.66 & 1067.16 & - & - & 875.76 \\ 
  M4 Quarterly & 622.57 & 574.34 & 573.19 & 604.51 & - & - & 610.51 \\ 
  M4 Monthly & 625.24 & 563.58 & 582.60 & 575.36 & - & - & 596.19 \\ 
  M4 Weekly & 336.82 & 333.32 & - & - & 296.15 & 321.61 & 293.21 \\ 
  M4 Daily & 178.27 & 178.86 & 193.26 & 179.67 & - & - & 181.92 \\ 
  M4 Hourly & 1218.06 & 1220.97 & - & - & 386.27 & 1310.85 & 257.39 \\ 
  Carparts & 0.55 & 0.53 & 0.56 & 0.56 & - & - & 0.41 \\ 
  Hospital & 21.76 & 18.54 & 17.97 & 19.60 & - & - & 19.24 \\ 
 \hline
\end{tabular}
\caption{Mean MAE Results}
\label{tab:mean_mae}
\end{center}
\end{table}

\begin{table}
\begin{center}
\footnotesize
\begin{tabular}{lccccccc}
\hline
\textbf{Dataset} & \textbf{SES} & \textbf{Theta} & \textbf{ETS} & \textbf{ARIMA} & \textbf{TBATS} & \textbf{DHR-ARIMA} & \textbf{PR} \\ 
\hline
NN5 Daily & 5.94 & 3.55 & 3.48 & 3.85 & - & - & 5.06 \\ 
  NN5 Weekly & 14.18 & 13.90 & - & - & 13.73 & 14.82 & 12.84 \\ 
  CIF 2016 & 107.09 & 103.39 & 70.43 & 80.66 & - & - & 95.13 \\ 
  Kaggle Daily & 51.05 & 51.64 & 69.27 & 46.27 & - & - & - \\ 
  Tourism Yearly & 4312.77 & 4085.98 & 4271.06 & 4623.59 & - & - & 4340.90 \\ 
  Tourism Quarterly & 1921.00 & 1114.30 & 1003.24 & 1047.01 & - & - & 992.12 \\ 
  Tourism Monthly & 967.57 & 478.45 & 457.04 & 462.53 & - & - & 474.72 \\ 
  Traffic Hourly & 0.02 & 0.02 & - & - & 0.02 & 0.03 & 0.02 \\ 
  Electricity Hourly & 210.20 & 210.20 & - & - & 127.05 & 215.60 & 137.88 \\ 
  M3 Yearly & 703.33 & 660.49 & 641.07 & 701.32 & - & - & 711.86 \\ 
  M3 Quarterly & 371.95 & 294.16 & 304.53 & 333.74 & - & - & 325.44 \\ 
  M3 Monthly & 517.09 & 420.80 & 408.92 & 412.47 & - & - & 479.18 \\ 
  M4 Yearly & 529.96 & 428.94 & 427.24 & 493.19 & - & - & 456.65 \\ 
  M4 Quarterly & 318.93 & 274.24 & 250.82 & 262.40 & - & - & 295.64 \\ 
  M4 Monthly & 291.89 & 249.73 & 244.21 & 243.12 & - & - & 280.83 \\ 
  M4 Weekly & 219.63 & 210.47 & - & - & 163.68 & 188.39 & 176.01 \\ 
  M4 Daily & 92.14 & 91.85 & 92.16 & 92.18 & - & - & 92.28 \\ 
  M4 Hourly & 49.20 & 49.21 & - & - & 33.77 & 30.75 & 14.21 \\ 
  Carparts & 0.33 & 0.25 & 0.33 & 0.33 & - & - & 0.25 \\ 
  Hospital & 6.67 & 6.67 & 6.67 & 6.83 & - & - & 6.67 \\ 
\hline
\end{tabular}
\caption{Median MAE Results}
\label{tab:median_mae}
\end{center}
\end{table}

\begin{table}
\begin{center}
\footnotesize
\begin{tabular}{lccccccc}
\hline
\textbf{Dataset} & \textbf{SES} & \textbf{Theta} & \textbf{ETS} & \textbf{ARIMA} & \textbf{TBATS} & \textbf{DHR-ARIMA} & \textbf{PR} \\ 
\hline
NN5 Daily & 8.23 & 5.28 & 5.22 & 6.05 & - & - & 7.26 \\ 
  NN5 Weekly & 18.82 & 18.65 & - & - & 18.53 & 18.55 & 18.62 \\ 
  CIF 2016 & 657112.42 & 804654.19 & 722397.37 & 526395.02 & - & - & 648890.31 \\ 
  Kaggle Daily & 590.11 & 583.32 & 650.43 & 595.43 & - & - & - \\ 
  Tourism Yearly & 106665.20 & 99914.21 & 104700.51 & 106082.60 & - & - & 89645.61 \\ 
  Tourism Quarterly & 17270.57 & 9254.63 & 10812.34 & 12564.77 & - & - & 11746.85 \\ 
  Tourism Monthly & 7039.35 & 2701.96 & 2542.96 & 3132.40 & - & - & 2739.43 \\ 
  Traffic Hourly & 0.04 & 0.04 & - & - & 0.05 & 0.04 & 0.03 \\ 
  Electricity Hourly & 1026.29 & 1026.36 & - & - & 743.35 & 1082.44 & 689.85 \\ 
  M3 Yearly & 1172.85 & 1106.05 & 1189.21 & 1662.17 & - & - & 1181.81 \\ 
  M3 Quarterly & 670.56 & 567.70 & 598.73 & 650.76 & - & - & 605.50 \\ 
  M3 Monthly & 893.88 & 753.99 & 755.26 & 790.76 & - & - & 830.04 \\ 
  M4 Yearly & 1154.49 & 1020.48 & 1052.12 & 1230.35 & - & - & 1000.18 \\ 
  M4 Quarterly & 732.82 & 673.15 & 674.27 & 709.99 & - & - & 711.93 \\ 
  M4 Monthly & 755.45 & 683.72 & 705.70 & 702.06 & - & - & 720.46 \\ 
  M4 Weekly & 412.60 & 405.17 & - & - & 356.74 & 386.30 & 350.29 \\ 
  M4 Daily & 209.75 & 210.37 & 229.97 & 212.64 & - & - & 213.01 \\ 
  M4 Hourly & 1476.81 & 1483.70 & - & - & 469.87 & 1563.05 & 312.98 \\ 
  Carparts & 0.78 & 0.78 & 0.80 & 0.81 & - & - & 0.73 \\ 
  Hospital & 26.55 & 22.59 & 22.02 & 23.68 & - & - & 23.48 \\ 
 \hline
\end{tabular}
\caption{Mean RMSE Results}
\label{tab:mean_rmse}
\end{center}
\end{table}

\begin{table}
\begin{center}
\footnotesize
\begin{tabular}{lccccccc}
\hline
\textbf{Dataset} & \textbf{SES} & \textbf{Theta} & \textbf{ETS} & \textbf{ARIMA} & \textbf{TBATS} & \textbf{DHR-ARIMA} & \textbf{PR} \\ 
\hline
NN5 Daily & 7.46 & 4.95 & 4.86 & 5.42 & - & - & 6.80 \\ 
  NN5 Weekly & 17.52 & 16.82 & - & - & 16.99 & 17.49 & 16.26 \\ 
  CIF 2016 & 129.06 & 118.29 & 85.77 & 103.14 & - & - & 109.09 \\ 
  Kaggle Daily & 74.58 & 75.16 & 98.97 & 68.13 & - & - & - \\ 
  Tourism Yearly & 4718.37 & 4615.95 & 4626.74 & 5174.76 & - & - & 4717.10 \\ 
  Tourism Quarterly & 2295.67 & 1392.89 & 1207.24 & 1196.05 & - & - & 1184.48 \\ 
  Tourism Monthly & 1250.26 & 675.10 & 598.88 & 603.66 & - & - & 596.26 \\ 
  Traffic Hourly & 0.03 & 0.03 & - & - & 0.03 & 0.04 & 0.02 \\ 
  Electricity Hourly & 256.22 & 256.22 & - & - & 181.79 & 275.52 & 171.57 \\ 
  M3 Yearly & 803.71 & 740.10 & 758.62 & 814.68 & - & - & 824.55 \\ 
  M3 Quarterly & 436.25 & 355.79 & 368.91 & 405.87 & - & - & 378.31 \\ 
  M3 Monthly & 633.56 & 516.79 & 495.97 & 497.97 & - & - & 582.04 \\ 
  M4 Yearly & 610.38 & 497.80 & 494.90 & 567.70 & - & - & 525.42 \\ 
  M4 Quarterly & 378.29 & 322.60 & 297.17 & 310.08 & - & - & 346.99 \\ 
  M4 Monthly & 348.59 & 299.02 & 293.25 & 292.51 & - & - & 333.30 \\ 
  M4 Weekly & 262.04 & 242.14 & - & - & 197.26 & 224.55 & 223.12 \\ 
  M4 Daily & 108.04 & 108.55 & 108.77 & 108.40 & - & - & 108.48 \\ 
  M4 Hourly & 61.40 & 61.58 & - & - & 42.90 & 42.93 & 19.89 \\ 
  Carparts & 0.71 & 0.65 & 0.71 & 0.71 & - & - & 0.58 \\ 
  Hospital & 8.26 & 8.20 & 8.25 & 8.45 & - & - & 8.25 \\ 
 \hline
\end{tabular}
\caption{Median RMSE Results}
\label{tab:median_rmse}
\end{center}
\end{table}

The globally trained PR models show a mixed performance compared with the traditional univariate forecasting models. The performance of the PR models is considerably affected by the number of past lags used during model training, performing better as the number of lags is increased. The number of lags we use during model training is quite high with the high-frequency datasets such as hourly, compared with the other datasets and hence, PR models generally show a better performance than the traditional univariate forecasting models on all error metrics across those datasets. But on the other hand, the memory and computational requirements are also increased when training PR models with large numbers of lags. Furthermore, the PR models show a better performance across intermittent datasets such as car parts, compared with the traditional univariate forecasting models. 

We note that the MASE values of the baselines are generally high on multi-seasonal datasets. For multi-seasonal datasets, we consider longer forecasting horizons corresponding to one week unless they are competition datasets. As benchmark in the MASE calculations, we use a seasonal na\"ive forecast for the daily seasonality. As therewith the MASE compares the forecasts of longer horizons (up to one week) with the in-sample sna\"ive forecasts obtained with shorter horizons (one day), the MASE values of multi-seasonal datasets are considerably greater than one across all baselines. Furthermore, the error measures are not directly comparable across datasets as we consider different forecasting horizons with different datasets.

\section{Conclusion}
\label{sec:conclusion}

Many businesses and industries nowadays rely on large quantities of time series from similar sources. Recently, global forecasting models have shown huge potential in providing accurate forecasts for such collections of time series compared with the traditional univariate benchmarks. However, there are currently no comprehensive time series forecasting benchmark data archives available that contain datasets to facilitate the evaluation of the new global forecasting algorithms. In this paper, we have presented the details of an archive that contains 20 publicly available time series datasets with different frequencies from varied domains. 
In addition to the datasets that are sets of series, which is the main focus of the archive, we furthermore include 6 datasets that contain single but very long series.

We have also characterised the datasets and have identified the similarities and differences among them by conducting a feature analysis exercise using tsfeatures and catch22 features extracted from each series. Finally, we have evaluated the performance of seven baseline forecasting models including six traditional univariate forecasting models: SES, Theta, ETS, ARIMA, TBATS, DHR-ARIMA, and a global forecasting model, PR, over all datasets across eight error metrics to enable other researchers to benchmark their own forecasting algorithms directly against those.

\bibliographystyle{elsarticle-harv}
\bibliography{sample}

\end{document}